\documentclass[lettersize,journal]{IEEEtran}
\usepackage{amsmath,amsfonts}
\usepackage{algorithmic}
\usepackage{algorithm}
\usepackage{array}
\usepackage[caption=false,font=normalsize,labelfont=sf,textfont=sf]{subfig}
\usepackage{textcomp}
\usepackage{stfloats}
\usepackage{url}
\usepackage{verbatim}
\usepackage{graphicx}
\usepackage{cite}
\usepackage{amsthm,amssymb,bm,mathrsfs}
\usepackage{graphicx,multicol}
\usepackage{multirow,booktabs,threeparttable}
\usepackage{color}
\usepackage{tabularx}
\usepackage{makecell}

\newcolumntype{Y}{>{\centering\arraybackslash}X}
\newcolumntype{L}{>{\raggedright\arraybackslash}X}

\graphicspath{{fig/}}

\usepackage{setspace}
\theoremstyle{remark}

\theoremstyle{remark}
\newtheorem*{remark}{Remark}

\begin{document}

\title{MERLOT: A Distilled LLM-based Mixture-of-Experts Framework for Scalable Encrypted Traffic Classification}

\author{Yuxuan Chen, Rongpeng Li, Zhifeng Zhao, and Honggang Zhang
        \thanks{Y. Chen and R. Li are with Zhejiang University, Hangzhou 310027, China, (email: \{cyx00, lirongpeng\}@zju.edu.cn).}

        \thanks{Z. Zhao is with Zhejiang Lab, Hangzhou 310012, China, as well as Zhejiang University, Hangzhou 310027, China (email: zhaozf@zhejianglab.com).}

        \thanks{H. Zhang is with City University of Macau, Macau, China (email: hgzhang@cityu.edu.mo).}}

\markboth{Journal of \LaTeX\ Class Files,~Vol.~14, No.~8, August~2021}%
{Shell \MakeLowercase{\textit{et al.}}: A Sample Article Using IEEEtran.cls for IEEE Journals}


\maketitle

\begin{abstract}
    We present \textbf{MERLOT}, a scalable \underline{m}ixture-of-\underline{e}xpert (MoE) based \underline{r}efinement of distilled \underline{l}arge language model \underline{o}ptimized for encrypted \underline{t}raffic classification. By applying model distillation techniques in a teacher-student paradigm, compact models derived from GPT-2-base retain high classification accuracy while minimizing computational costs. These models function as specialized experts in an MoE architecture, dynamically assigned via a gating network. Unlike generation-based methods, our approach directly classifies encrypted traffic using the final decoder token with contextual feature embedding as input. Experiments on $10$ datasets show superior or competitive performance over the state-of-the-art models while significantly reducing resource demands, underscoring its effectiveness and robustness.
\end{abstract}


\section{Introduction}
As an essential function in modern networks, traffic classification lays the very foundation for ensuring real-time monitoring, optimization, and security  \cite{zhao2021network}. However, the widespread adoption of encryption and the increasing complexity of application behaviors have rendered traditional techniques—such as port-based analysis and deep packet inspection \cite{cascarano2011optimizing}—ineffective. To address these challenges, data-driven approaches leveraging machine learning (ML) and deep learning (DL) \cite{yang2021deep} have emerged with the promising ability to automate feature extraction and adapt to diverse, dynamic traffic patterns \cite{azab2024network}. %These methods form the foundation for advancing traffic classification in complex, encrypted environments.
%Recent advancements in deep learning have brought 
Furthermore, motivated by the success of large language models (LLMs) \cite{radford2019language}, $110$-million-parameter ET-BERT \cite{lin2022bert}, $117$-million-parameter NetGPT \cite{meng2023netgpt} and $7$-billion-parameter Llama2-based TrafficLLM \cite{ZGC-LLM-Safety_TrafficLLM_2024} demonstrate astonishing %into the realm of network traffic analysis, demonstrating their 
ability to capture the intricate traffic features through self-supervised learning. However, the huge computational and memory requirements in TrafficLLM \cite{ZGC-LLM-Safety_TrafficLLM_2024} often make it impractical for real-time deployment, especially in resource-limited network entities \cite{ZGC-LLM-Safety_TrafficLLM_2024}; while the other alternatives \cite{lin2022bert,meng2023netgpt} encounter degraded performance. %Nevertheless, their application in network traffic classification remains limited. Existing solutions frequently utilize monolithic, large-scale LLMs, which impose significant computational and memory burdens \cite{ZGC-LLM-Safety_TrafficLLM_2024}. 
Meanwhile, these models, while effective in generalized scenarios, lack the efficiency and specialization required for diverse traffic classification tasks \cite{wu2024netllm}. Moreover, their over-reliance on prompt-based generative workflows introduces additional latency and inefficiency, underscoring the need for innovative methods that achieve a balance between accuracy and resource efficiency.

To address these challenges, we propose MERLOT, a scalable \underline{m}ixture-of-\underline{e}xpert (MoE) based \underline{r}efinement of distilled \underline{l}arge language model \underline{o}ptimized for encrypted \underline{t}raffic classification. At its core, MERLOT adopts a Mixture-of-Experts (MoE) architecture \cite{jacobs1991adaptive} to ensure scalability and respects the advantages of pretrained foundation models with GPT-2-base as the basis. To maintain high accuracy while drastically reducing computational demands, MERLOT further distills to obtain models from the GPT-2-base using a teacher-student paradigm \cite{hinton2015distilling}. Specifically, MERLOT leverages task-specific teacher models, fine-tuned on individual datasets, to generate soft labels guiding the training of student models, which (due to its significantly smaller size) minimizes resource consumption while retaining classification efficacy. Afterward, the distilled models are integrated through the MoE architecture, where a gating mechanism dynamically assigns each traffic classification instance to the most relevant expert model. Such provisional expert activation contributes to requiring shrunk %This dynamic allocation ensures optimal resource utilization by activating only one expert model per input instance, significantly reducing 
computational overhead \cite{jacobs1991adaptive} than monolithic approaches \cite{lin2022bert,ZGC-LLM-Safety_TrafficLLM_2024}. Furthermore, MERLOT departs from traditional generative classification methods \cite{ghanavi2021generative, meng2023netgpt}, which are contingent on task-specific prompt-based workflows, and directly utilizes the final token in the decoder to aggregate sequential information for direct classification. %This approach leverages the autoregressive nature of decoder-only LLMs, where the final token inherently captures global dependencies within the input sequence. 
By avoiding the overhead of constructing and interpreting prompts, this method streamlines the classification process. To further enhance performance, we augment the input data representation by embedding key metadata, such as protocol types and IP addresses, within concise natural language prompts to preserve essential semantics. %This augmentation activates the model’s inherent contextual understanding, improving interpretability and accuracy.
Eventually, the $0.66$-billion-parameter MERLOT yields competitive or even superior performance than $7$-billion-parameter TrafficLLM \cite{ZGC-LLM-Safety_TrafficLLM_2024}, but consumes $85-90\%$ less inference time and memory usage%Experiments conducted on $10$ benchmark datasets demonstrate competitive or superior classification performance in eight tasks
, affirming the framework’s balance of efficiency and accuracy.

The remainder of this letter is organized as follows. Section \ref{sec:problem} formulates the encrypted traffic classification problem, while Section \ref{sec:framework} provides the in-depth description of MERLOT. Section \ref{sec:results} gives extensive experimental results. Section \ref{sec:conclusion} concludes the letter.

%In summary, LLMoE-Traffic addresses critical challenges in network traffic classification by balancing high accuracy with computational efficiency, offering a viable solution for real-time applications in resource-constrained environments. This letter highlights the framework’s design, theoretical underpinnings, and empirical performance, establishing its potential as an innovative step forward in traffic classification methodologies.

\section{Problem Formulation}
\label{sec:problem}

This letter aims to identify the traffic category from encrypted traffic logs. Notably, due to obscured semantic content of the payload, encryption leaves only statistical patterns in metadata, such as flow timing, packet lengths, and protocol headers. 
%This creates a problem of high-dimensional, low-signal feature representation, where the task is to identify subtle patterns and correlations that distinguish between classes, even in the presence of noise or overlapping feature distributions.
%The classification of network traffic in contemporary communication systems is increasingly complicated by the widespread adoption of encryption protocols. Encryption, while essential for ensuring data confidentiality and security, transforms raw packet payloads into ciphertext, rendering traditional payload-based traffic analysis techniques ineffective. As a result, encrypted traffic primarily exposes high-level metadata while obscuring the semantic content of the payload. This constraint fundamentally alters the nature of the classification problem, shifting the focus from payload inspection to the inference of traffic patterns based on observable, yet limited, features.
Formally, the set of encrypted network traffic metadata can be represented as $\mathcal{X} = \{\bm{x}_1, \bm{x}_2, \dots, \bm{x}_N\}$, where each $\bm{x}_i$ is a feature vector (i.e., traffic instance) comprising the aforementioned observable characteristics of a packet or flow. These features lack direct semantic context due to encryption but are often implicitly correlated with the underlying traffic class $y_i \in \mathcal{Y}$, where \(\mathcal{Y}\) is the set of predefined traffic categories (e.g., video streaming, VoIP, malicious botnet traffic). For example, the combination of source and destination IP addresses, along with temporal flow characteristics, may indicate specific application behaviors or usage patterns. Therefore, the goal of the classification task is to recover the mapping $f: \mathcal{X} \to \mathcal{Y}$ that assigns each instance $\bm{x}_i$ to its corresponding class $y_i$, despite the absence of payload-level information.

In this context, the classification problem can be mathematically formalized as minimizing a supervised loss function
\begin{align}
\mathcal{L} = \frac{1}{N} \sum\nolimits_{i=1}^N \mathcal{L}(f(\bm{x}_i;\theta), y_i),
\end{align}
where \(\mathcal{L}(\cdot)\) denotes the classification loss, which commonly uses a cross-entropy one, and $f(\cdot;\theta)$ represents the model parameterized by \(\theta\). %For encrypted traffic, this optimization must account for the limited feature space $\bm{x}_i$, where traditional content-based signal information is replaced with statistical patterns. Effective classification under these constraints requires models that can leverage subtle dependencies in the metadata while generalizing across heterogeneous traffic scenarios.
%A critical theoretical insight is that metadata features often exhibit implicit correlations with the underlying traffic class, despite the obfuscation caused by encryption. For example, the combination of source and destination IP addresses, along with temporal flow characteristics, may indicate specific application behaviors or usage patterns. Extracting and encoding these correlations into meaningful feature representations is essential for recovering traffic classes from encrypted flows. Thus
In other words, the task of encrypted traffic classification is fundamentally one of pattern recognition in a reduced-information domain, requiring advanced models capable of capturing latent relationships and leveraging indirect features in the metadata.

\section{The Proposed MERLOT Framework}
\label{sec:framework}
\begin{figure}
	\centering
	\includegraphics[width=0.495\textwidth]{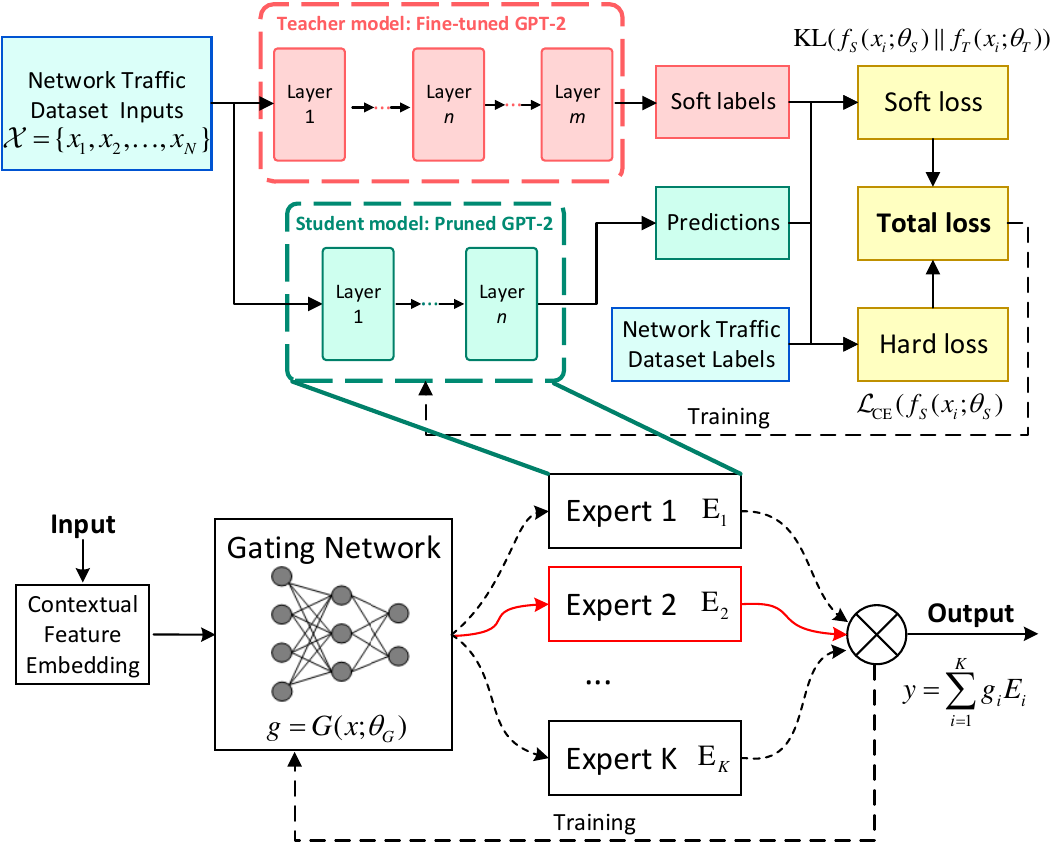}
	\caption{Overview of the MERLOT architecture.}
    \vspace{-1em}
	\label{fig:system_model}
\end{figure}

%The motivation for the proposed LLMoE-Traffic framework stems from the escalating complexity of network traffic, particularly in encrypted environments, where traditional methods fall short in scalability, efficiency, and precision. Addressing these challenges, our framework combines model distillation with an MoE architecture, enabling adaptive traffic classification while minimizing computational demands. This design achieves an optimal balance between resource efficiency and classification accuracy, making it particularly suitable for deployment in resource-constrained scenarios. 
Overall, Fig. \ref{fig:system_model} illustrates the framework of MERLOT.
In brief, MERLOT introduces a novel framework that capably addresses the computational inefficiencies and scalability limitations of existing approaches \cite{lin2022bert,ZGC-LLM-Safety_TrafficLLM_2024}. By leveraging model distillation \cite{hinton2015distilling}, dynamic expert selection \cite{jacobs1991adaptive}, and augmented input representations, it provides a theoretically grounded and practically efficient solution to encrypted network traffic classification challenges.
\subsection{Foundation Model}
%GPT-2-base 描述
The proposed MERLOT framework is based on GPT-2-base \cite{radford2019language}, a decoder-only LLM. % renowned for its autoregressive capabilities and efficiency in handling sequential data. The model is designed to predict the next token in a sequence by leveraging a self-attention mechanism, which captures dependencies across the input sequence. 
With $12$ transformer layers, a hidden dimension of $768$, and approximately $117$-million parameters, GPT-2-base benefits from 
%achieves a balance between computational efficiency and representational power, making it a compelling choice for resource-constrained scenarios.
%The architecture's 
the autoregressive nature that each token progressively accumulates information from its predecessors, enabling the final token in the sequence to aggregate global context. This characteristic aligns with the requirements of network traffic classification, where the input is represented as sequential metadata features. Furthermore, %GPT-2-base's architecture is well-suited for deployment within the MoE framework. 
its moderate parameter count and computational demands allow for effective distillation into lightweight, task-specific models.
%, enabling the dynamic allocation of computational resources while preserving classification accuracy. 
We denote fine-tuned GPT2-base as $f_T(\bm{x}; \theta_T) $.
\subsection{Model Distillation}
Although GPT-2-base is smaller than many mainstream LLMs, its resource requirements remain prohibitive in scenarios where multiple models are deployed concurrently for specialized traffic classification tasks. Therefore, we employ model distillation to transform the GPT-2-base model into a collection of lightweight, task-specific models.
%To meet the computational and deployment demands of real-time network traffic classification, we employ model distillation to transform the GPT-2-base model into a collection of lightweight, task-specific models. Network edge environments, characterized by limited GPU resources, restricted memory capacity, and strict low-latency requirements, impose significant constraints on model size and inference efficiency. 
In particular, model distillation effectively compresses the knowledge of the ``teacher" model $f_T(\bm{x}; \theta_T) $ into a smaller ``student" model $f_S(\bm{x}; \theta_S) $. %, allowing us to preserve the essential functionalities of the original model while substantially reducing its computation requirement. 
Correspondingly, the distillation process minimizes a composite loss function defined as
\begin{align}
\mathcal{L}_{\text{total}} &= \frac{1}{|\mathcal{B}|} \sum_{i \in \mathcal{B}} \Big[(1 - \alpha) \cdot \underbrace{\mathcal{L}(f_S(\bm{x}_i; \theta_S), y_i)}_{\text{Hard Loss}} \notag \\
&+ \alpha \cdot \underbrace{\text{KL}(f_S(\bm{x}_i; \theta_S) \parallel f_T(\bm{x}_i; \theta_T))}_{\text{Soft Loss}}\Big],
\label{eq:distillation loss}
\end{align}
where the (cross-entropy) loss $\mathcal{L}$ in a batch $\mathcal{B}$ ensures that the student model aligns with true labels, and $\text{KL}$ denotes the Kullback-Leibler divergence, encouraging the student model to match the output distribution of the teacher model. Through the composite loss, the student model learns from both hard labels (i.e., direct classifications) and soft labels (i.e., probabilistic outputs from the teacher model), balancing accuracy with computational efficiency. By adjusting $\alpha$, we control the relative emphasis on learning directly from true labels versus the teacher’s confidence distribution, which encapsulates inter-class relationships and relative uncertainty levels.

\begin{remark}
The advantages of using model distillation stem from both its theoretical and practical benefits for traffic classification tasks. Distillation enables the student model to inherit a ``richer" representation of the input data, as soft labels provide information beyond binary class distinctions, capturing nuances in how classes relate to one another. This capability is particularly valuable in network traffic analysis, where subtle variations between traffic classes—such as benign versus suspicious traffic or different types of encrypted flows—possibly lead to distinguishable outcomes.%require more than binary or discrete feature representations. Owing to the guidance of the teacher model, the student model, therefore, attains a nuanced, probabilistically-informed understanding of traffic types, improving its ability to generalize to complex classification scenarios.
\end{remark}
\subsection{Dynamic Expert Selection}
%The MoE architecture dynamically routes each traffic instance to a specialized expert model. This adaptive routing is facilitated by a gating network, implemented as a distilled version of GPT-2, which leverages its deep representational capacity to analyze input features and make expert selections. 
The MoE architecture addresses the heterogeneity of collected network traffic by allocating distinct computational resources to specific traffic types, ensuring both precision and computational efficiency. Without loss of generality, assume there exist $K$ expert models and $ \{E_1, E_2, \dots, E_K\} $ denote the corresponding available expert models, where each expert $ E_i $ is optimized for a specific subtask or traffic classification type. For an incoming traffic instance $ \bm{x} $, the gating function $ G(\bm{x}; \theta_G) $, parameterized by $ \theta_G $ and implemented as a distilled, pruned GPT-2-base model, evaluates the input $ \bm{x} $ and outputs a binary selection vector $ \bm{g}  = \{g_1, \cdots, g_K\} \in \{0, 1\}^K $. Notably, $ g_i = 1 $ if and only if the expert $ E_i $ is chosen for processing $ \bm{x} $. Mathematically,
\begin{align}
\bm{g} = G(\bm{x}; \theta_G), \quad \text{s.t.} \quad g_i \in \{0, 1\}, \, \sum\nolimits_{i=1}^K g_i = 1,
\end{align}
where the binary constraint on $ \bm{g} $ ensures that exactly one expert is activated for each input, minimizing unnecessary computational overhead. The selected expert $ E_i $ is then applied to produce the classification output $ y $ through
\begin{align}
y = \sum\nolimits_{i=1}^K g_i E_i(\bm{x}; \theta_{E_i}),
\label{eq:MoE}
\end{align}
where $ E_i(\bm{x}; \theta_{E_i}) $ denotes the output of expert $ E_i $ with parameters $ \theta_{E_i} $. Since $ \bm{g} $ is a one-hot vector, only the chosen expert contributes to the final output, enabling precise task-specific inference. % while conserving resources.

\begin{remark}
In network traffic classification, different types of traffic data—such as encrypted flows or specific attack signatures—exhibit distinct feature patterns that often require specialized processing. Our experimental experience indicates that a single  unified model might struggle to capture the full diversity of these patterns without sacrificing efficiency or interpretability. 
The MoE framework, by contrast, decomposes the task into sub-problems, allowing each expert to focus on a specific aspect of the whole traffic classification spectrum. Furthermore, the MoE setup in Eq. \eqref{eq:MoE} with a hard gating mechanism provides deterministic routing \cite{jacobs1991adaptive}, which not only simplifies interpretability but also ensures consistent performance by preventing fluctuating activation patterns that may arise in soft gating systems \cite{zhou2022mixture, huang2024harder}. In a nutshell, this modularity not only enhances classification accuracy by matching inputs to the most suitable expert but also mitigates the computational burden, as only the most relevant sub-model is active at any given time. %This property is especially beneficial in real-time, resource-constrained environments such as edge networks, where stability and predictability in model behavior are critical.
\end{remark}

\begin{table}[tbp]
\centering
\caption{Description of encrypted network traffic datasets.}
\setlength{\tabcolsep}{1pt}
%\renewcommand{\arraystretch}{1.5} 
%\resizebox{\linewidth}{!}{ 
    \begin{tabular}{@{}lm{3.5cm}ll@{}}
    \toprule
    \textbf{Dataset}       & \textbf{Task}                          & \textbf{\#Samples}   & \textbf{\#Label}\\ \midrule
    APP-53 2023            & Concept Drift Analysis    & 109.8K             & 54 \\
    CSIC 2010              & Web Attack Detection             & 34.5K              & 2 \\
    CSTNET 2023            & Encrypted App Classification          & 97.6K              & 20 \\
    CW-100 2018            & Website Fingerprinting     & 7.4K               & 100 \\
    DAPT 2020              & APT Attack Detection    & 10.0K              & 2 \\
    DoHBrw 2020            & Malicious DoH Detection         & 47.8K              & 2 \\
    ISCX Botnet 2014       & Botnet Detection        & 25.0K              & 5 \\
    ISCX Tor 2016          & Tor Behavior Detection               & 40.0K              & 8 \\
    ISCX VPN 2016          & Encrypted VPN Detection            & 64.8K              & 14 \\
    USTC TFC 2016          & Malware Traffic Detection              & 50.7K              & 20 \\ \bottomrule
    \end{tabular}
%}
\label{tab:datasets}
% \vspace{-1em}
\end{table}
\begin{figure}
	\centering
	\includegraphics[width=0.485\textwidth]{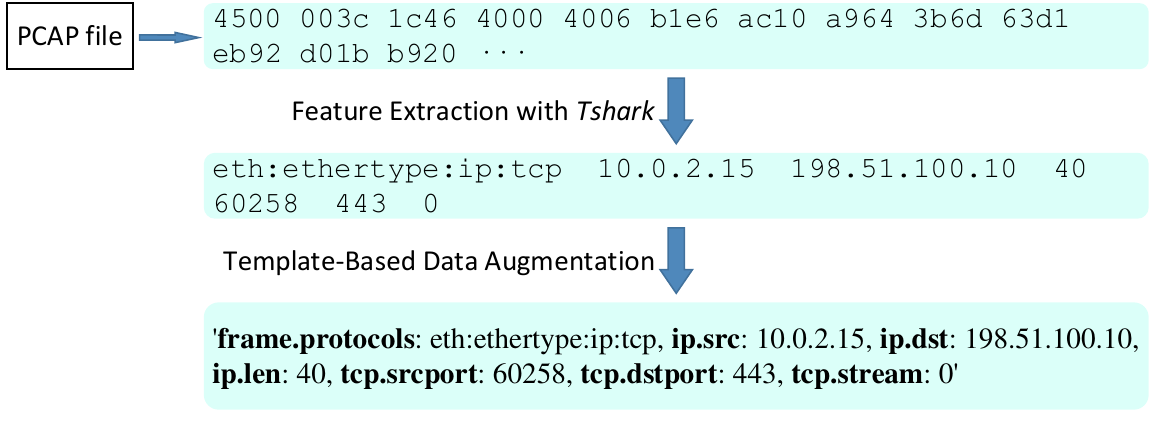}
	\caption{Example of contextual feature embedding.}
	\label{fig:data_argument}
    \vspace{-1em}
\end{figure}

\begin{table*}[htbp]
    \centering
    \caption{Performance across $10$ network traffic classification datasets.}
    \label{tab:comparison_combined} 
    \small
    \setlength{\tabcolsep}{1.8pt}
    \begin{minipage}{\textwidth}
        \centering
        %\resizebox{\textwidth}{!}{ % Resize to fit the text width
            \begin{tabular}{l|ccc|ccc|ccc|ccc|ccc}
                \toprule
                Dataset & \multicolumn{3}{c|}{ISCX Tor 2016} & \multicolumn{3}{c|}{ISCX VPN 2016} & \multicolumn{3}{c|}{APP-53 2023} & \multicolumn{3}{c|}{CSTNET 2023} & \multicolumn{3}{c}{CW-100 2018} \\ \hline
                Method  & PR    & RC    & F1    & PR    & RC    & F1    & PR    & RC    & F1    & PR    & RC    & F1    & PR    & RC    & F1    \\ \hline
                ET-BERT      & 0.9186 & 0.9430 & 0.9368 & 0.9567 & 0.9420 & 0.9539 & 0.8540 & 0.8494 & 0.8506 & 0.9581 & 0.9478 & 0.9496 & 0.8670 & 0.8650 & 0.8660  \\
                TrafficLLM      &  0.9801 & 0.9871 & 0.9810 & 0.9960 & 0.9970 & 0.9970 & 0.9325 & 0.9315 & 0.9320 & 0.9678 & 0.9369 & 0.9599 & 0.9370 & 0.9360 & 0.9366 \\
                MERLOT (500M)     & 0.9850 & 0.9844 & 0.9845 & 0.9920 & 0.9920 & 0.9920 & 0.8488 & 0.8420 & 0.8454 & 0.9986 & 0.9986 & 0.9986 & 0.8150 & 0.7794 & 0.7968 \\
                MERLOT (660M)     & 0.9850 & 0.9844 & 0.9845 & 0.9920 & 0.9920 & 0.9920 & 0.8622 & 0.8580 & 0.8601 & 0.9996 & 0.9996 & 0.9996 & 0.8693 & 0.8250 & 0.8466 \\
                MERLOT (1.25B)    & 0.9866 & 0.9850 & 0.9858 & 0.9923 & 0.9923 & 0.9923 & 0.8755 & 0.8650 & 0.8702 & 0.9998 & 0.9998 & 0.9998 & 0.9039  & 0.8750 & 0.8892 \\
                 \midrule
            \end{tabular}
        %}
    \end{minipage}

    \vspace{-0.1cm}

    \begin{minipage}{\textwidth}
        \centering
        %\resizebox{\textwidth}{!}{ 
            \begin{tabular}{l|ccc|ccc|ccc|ccc|ccc}
                \midrule
                Dataset & \multicolumn{3}{c|}{ISCX Botnet 2014} & \multicolumn{3}{c|}{USTC TFC 2016} & \multicolumn{3}{c|}{CIC DoHBrw 2020} & \multicolumn{3}{c|}{DAPT 2020} & \multicolumn{3}{c}{CSIC 2010} \\ \hline
                Method  & PR    & RC    & F1    & PR    & RC    & F1    & PR    & RC    & F1    & PR    & RC    & F1    & PR    & RC    & F1    \\ \hline
                ET-BERT          & 0.9503 & 0.9462 & 0.9489 & 0.9621 & 0.9508 & 0.9587 & 0.8927 & 0.8674 & 0.8467 & 0.9450 & 0.9423 & 0.9435 & 0.9021 & 0.8920 & 0.8995 \\
                TrafficLLM      & 0.9992 & 0.9992 & 0.9992 & 0.9950 & 0.9957 & 0.9950 & 0.9940 & 0.9940 & 0.9939 & 0.9820 & 0.9806 & 0.9810 & 0.9870 & 0.9823 & 0.9845 \\
                MERLOT (500M)     & 0.9984 & 0.9984 & 0.9984 & 0.9925 & 0.9925 & 0.9925 & 0.9999 & 0.9999 & 0.9999 & 0.9600 & 0.9610 & 0.9605 & 0.9607 & 0.9607 & 0.9607 \\
                MERLOT (660M)     & 0.9992 & 0.9992 & 0.9992 & 0.9953 & 0.9953 & 0.9953 & 0.9999 & 0.9999 & 0.9999 & 0.9640 & 0.9630 & 0.9635 & 0.9906 & 0.9906 & 0.9906\\
                MERLOT (1.25B)    & 0.9999 & 0.9999 & 0.9999 & 0.9977 & 0.9976 & 0.9976 & 0.9999 & 0.9999 & 0.9999 & 0.9651 & 0.9640 & 0.9645 & 0.9992 & 0.9992 & 0.9992\\ \bottomrule
            \end{tabular}
        %}
    \end{minipage}
% \vspace{-0.5em}
\end{table*}

% \begin{figure*}
% 	\centering
% 	\includegraphics[width=0.975\textwidth]{fig/figure_bar.pdf}
% 	\caption{Model performance comparison.}
% 	\label{fig:figure_bar}
% \end{figure*}

\subsection{Contextual Feature Embedding}
To enhance interpretability and robustness in traffic classification, our framework employs a contextual feature embedding strategy that augments each network traffic sample with essential metadata description in a natural language format. For example, key traffic attributes—such as protocol types, source IP addresses, and destination IP addresses—are incorporated as structured text prompts, thus better leveraging the model's language comprehension capabilities for nuanced classification. Hence, for each raw traffic sample, we perform feature extraction using tools such as Tshark \cite{wireshark} to obtain relevant metadata, including protocol layers, IP addresses, and port numbers. As depicted in Fig. \ref{fig:data_argument}, this metadata is then embedded within the input sequence as labeled descriptors, yielding a hybrid representation that aligns with the model's pretrained text processing strengths. % This structured augmentation allows the model to interpret network data with an enriched context, facilitating a deeper understanding of traffic patterns.

\begin{remark}
By embedding these metadata tokens, the model can more effectively capture inter-attribute relationships, improving its feature representation capacity and classification accuracy, particularly for complex, encrypted, and heterogeneous traffic classification scenarios. This technique enhances both interpretability and robustness in traffic classification tasks.
\end{remark}
\section{Experimental Settings and Performance Evaluation}
\label{sec:results}

\subsection{Experimental Setup}
We evaluate the performance of the MERLOT framework on $10$ benchmark network traffic datasets, including APP-53 2023, CSIS2010, CSTNET 2023, CW-100 2018, DAPT 2020, DoHBrw 2020, ISCX Botnet 2014, ISCX Tor 2016, ISCX VPN 2016, and USTC TFC 2016 \cite{ZGC-LLM-Safety_TrafficLLM_2024}. These datasets encompass a variety of traffic types—including encrypted, malicious, and legitimate flows—representing a diverse set of classification challenges. Table \ref{tab:datasets} provides an overview of these $10$ datasets,
For all datasets, the training and test sets are split as a 95:5 ratio, to ensure consistent evaluation.

Experiments are conducted on an NVIDIA A800 GPU to ensure efficient training and inference across both original and distilled models. All models are trained for 5 epochs, maintaining consistency across tasks for a fair performance comparison. In the distillation process, we set the distillation factor $\alpha = 0.5$ to balance the influence of true labels and teacher outputs, and a temperature parameter of $2.0$ to smooth the teacher’s output distribution, facilitating effective knowledge transfer to the student models. Finally, we assess the performance in terms of the commonly adopted precision (PR), recall (RC), and F1-Score (F1) \cite{vanRijsbergen1979}.
% following metrics.
% \begin{itemize}
%     \item Precision (PR) measures the proportion of correctly predicted positive instances out of all instances predicted as positive, namely,
%     \begin{align}
%     \text{Precision} = \frac{\text{True Positives (TP)}}{\text{True Positives (TP)} + \text{False Positives (FP)}}. \notag
%     \end{align}
%     \item Recall (RC), also referred to as Sensitivity or True Positive Rate, quantifies the proportion of actual positive instances correctly identified by the model, that is,
%     \begin{align}
%     \text{Recall} = \frac{\text{True Positives (TP)}}{\text{True Positives (TP)} + \text{False Negatives (FN)}}. \notag
%     \end{align}
%     \item F1-Score (F1) is the harmonic mean of Precision and Recall, providing a single metric to balance both aspects as
%     \begin{align}
%     \text{F1-Score} = 2 \cdot \frac{\text{Precision} \cdot \text{Recall}}{\text{Precision} + \text{Recall}}. \notag
%     \end{align}
% \end{itemize}
%These metrics collectively evaluate the model’s ability to classify network traffic accurately, balancing the trade-off between identifying true positives and minimizing false positives or negatives.

\subsection{Performance Evaluation}
To validate the effectiveness of the proposed MERLOT framework, we evaluate its classification performance across $10$ network traffic datasets, benchmarking against ET-BERT \cite{lin2022bert} and the state-of-the-art model TrafficLLM \cite{ZGC-LLM-Safety_TrafficLLM_2024}. 

As shown in Table \ref{tab:comparison_combined}, our framework achieves superior precision, recall, and F1 scores on $6$ of $10$ datasets than the $7$-billion-parameter TrafficLLM, while performing comparably on the remaining datasets. %This consistent performance demonstrates our method’s robustness, particularly in handling encrypted and obfuscated traffic, where specialized models are essential for high classification accuracy. The integration of model distillation and MoE architectures enables our framework to adapt to specific traffic patterns, leading to enhanced precision across diverse classification tasks.
% Figure \ref{fig:figure_bar} further illustrates performance across representative datasets, visually comparing the distilled student models with the teacher model and TrafficLLM. The bar charts highlight the competitiveness of our distilled models, which closely match or exceed the performance of the larger teacher model and generally outperform TrafficLLM, despite reduced computational complexity.
On the other hand, the $600$-million-parameter MERLOT framework employs an MoE architecture, comprising $10$ distilled expert models and a gating network, all derived from the GPT-2-base. Specifically, each expert model contains $3$ transformer layers with a hidden dimension of $768$, totaling approximately $85$-million parameters per expert, while the gating network has $935$-million parameters. In contrast, the LLaMA2-7B model in TrafficLLM \cite{ZGC-LLM-Safety_TrafficLLM_2024} has $32$ transformer layers with a hidden size of $4096$. 
Therefore, since the computational complexity of a transformer-based LLM can be approximated as $O(L \cdot L_{\text{seq}} \cdot D_{\text{hidden}}^2) $ \cite{vaswani2017attention}, our 3-layer MERLOT model leads to an $85-90\%$ decrease in inference time compared to TrafficLLM \cite{ZGC-LLM-Safety_TrafficLLM_2024}, corroborated by empirical results as well.
%In terms of computational complexity, the LLaMA2-7B model has a time complexity of $O(L \cdot L_{\text{seq}} \cdot D_{\text{hidden}}^2) $, where $L = 32 $, $D_{\text{hidden}} = 4096 $. Our 3-layer LLMoE-Traffic, by comparison, has a complexity of $O(L' \cdot L_{\text{seq}} \cdot D_{\text{hidden}}'^2) $, with $L' = 3 $ and $D_{\text{hidden}}' = 768 $, yielding an approximately 85-90\% reduction in inference time over LLaMA2-7B. %Furthermore, empirical results corroborate these findings, showing that our 3-layer LLMoE-Traffic achieves an $85-90\%$ decrease in inference time compared to TrafficLLM.
These results substantiate our framework’s adaptability and efficiency, balancing high classification accuracy with reduced resource consumption, supporting its suitability for deployment in resource-constrained environments.
\begin{figure}
	\centering
	\includegraphics[width=0.975\linewidth]{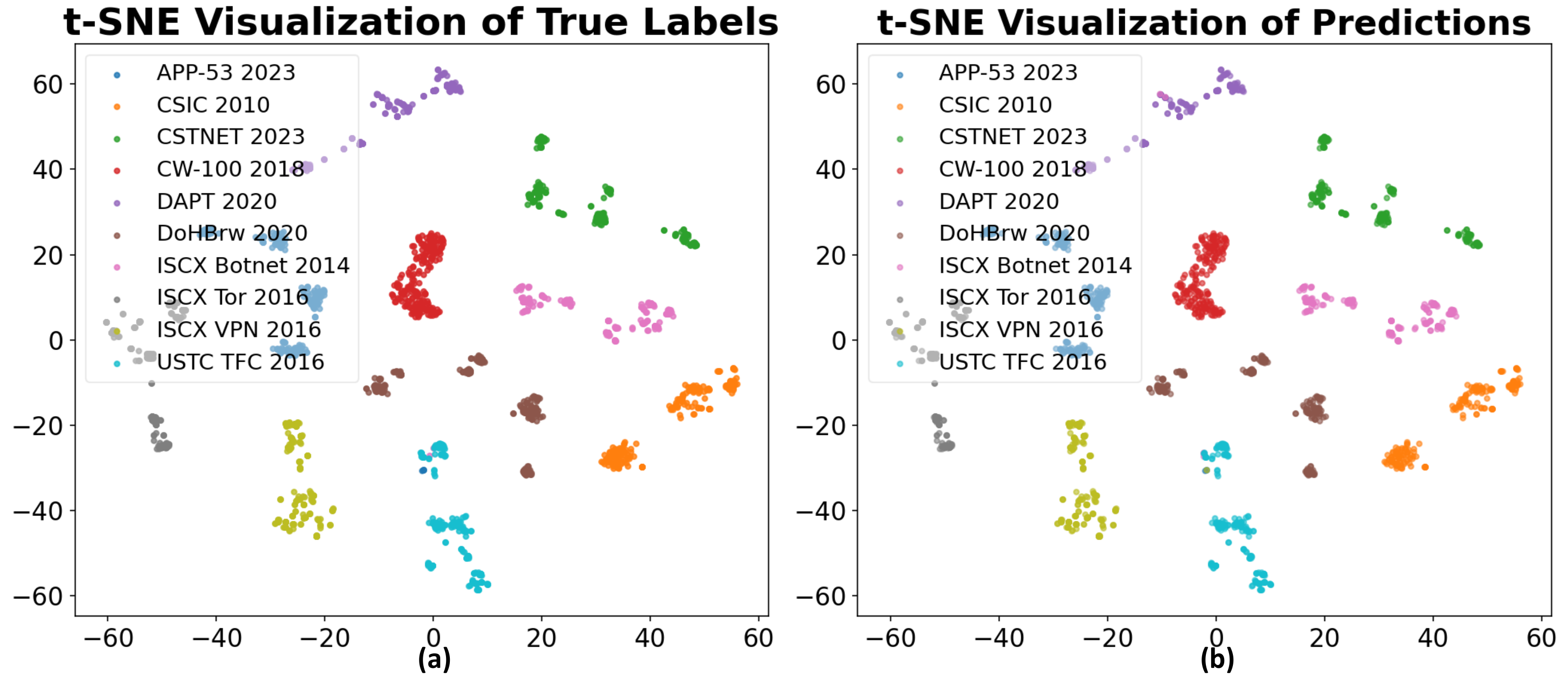}
	\caption{T-SNE visualizations of input embeddings and classification outputs. Left: True labels. Right: Predicted labels by the gating function and expert models in MoE.}
	\label{fig:figure_sne}
    \vspace{-1em}
\end{figure}

To further validate the effectiveness of the adopted MoE, we perform a t-SNE visualization \cite{van2008visualizing} of the input embedding features and compare the high-dimensional feature space to its classification outputs. It can be observed from the left panel of Fig. \ref{fig:figure_sne} that distinct clusters are formed for true labels from different traffic datasets. This indicates that the embedding features possess strong discriminative power in the high-dimensional space, providing a solid foundation for the deterministic gating in MoE. Meanwhile, the right panel of Fig. \ref{fig:figure_sne} shows the corresponding distribution of predicted labels generated by the gating mechanism and expert models. The high degree of consistency between the true and predicted labels demonstrates the classifier's capability to effectively identify data patterns. 

\begin{table*}[t]
    \centering
    \caption{Performance comparison with and without contextual feature embedding (CFE).}
        \begin{tabular}{l|ccc|ccc|ccc}
            \toprule
            Dataset & \multicolumn{3}{c|}{DAPT 2020} & \multicolumn{3}{c|}{CW-100 2018} & \multicolumn{3}{c}{USTC TFC 2016} \\ \hline
            Method  & PR    & RC    & F1    & PR    & RC    & F1    & PR    & RC    & F1    \\ \hline
            w/ CFE   & 0.9651 & 0.9640 & 0.9645 & 0.9039 & 0.8750 & 0.8892 & 0.9977 & 0.9976 & 0.9976 \\ 
            w/o CFE  & 0.9624 & 0.9600 & 0.9612 & 0.8967 & 0.8700 & 0.8831 & 0.9964 & 0.9965 & 0.9964 \\ 
            \bottomrule
        \end{tabular}\label{tab:data_augmentation_comparison}
\end{table*}
\begin{figure*}
	\centering
	\includegraphics[width=0.975\textwidth]{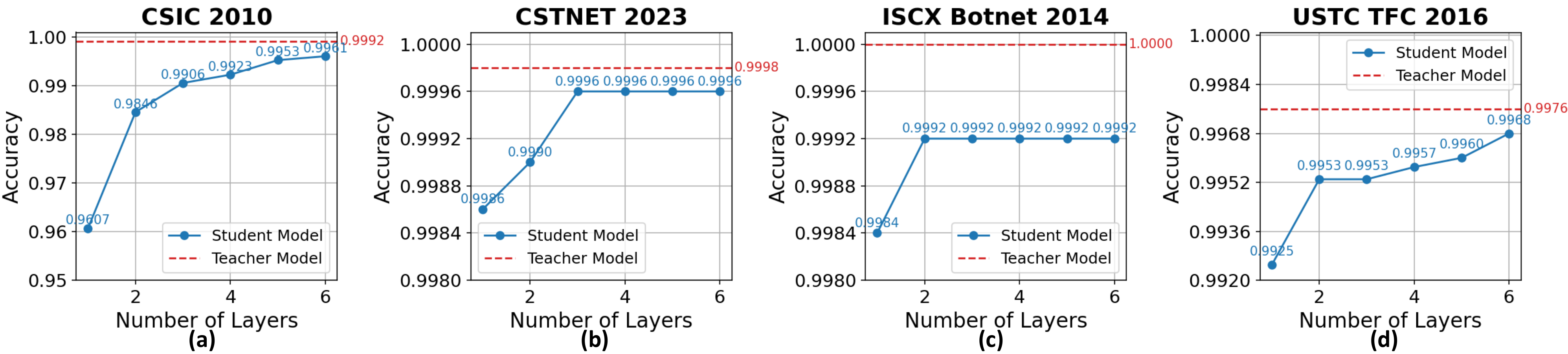}
	\caption{Performance variations with respect to different number of layers in the student model.}
	\label{fig:figure_layer}
    \vspace{-1em}
\end{figure*}

\begin{table}[tbp]
\centering
\caption{Effect of distillation hyperparameters on performance under USTC TFC 2016 dataset.}
\setlength{\tabcolsep}{6pt}
%\renewcommand{\arraystretch}{1.2}

% First sub-table: Effect of alpha
\textbf{(a) Effect of \(\alpha\) (Temperature \( = 2.0\))}
\begin{tabular}{@{}cccc@{}}
\toprule
\textbf{\(\alpha\)} & \textbf{Precision} & \textbf{Recall} & \textbf{F1-Score} \\ \midrule
% 0.3                 & 0.9934              & 0.9933           & 0.9933             \\
0.4                 & 0.9940              & 0.9940           & 0.9940             \\
0.5                 & 0.9953              & 0.9953           & 0.9953             \\
0.6                 & 0.9942              & 0.9941           & 0.9941             \\
% 0.7                 & 0.9929              & 0.9928           & 0.9928             \\ 
\bottomrule
\end{tabular}

\vspace{0.3cm} % Space between sub-tables

% Second sub-table: Effect of temperature
\textbf{(b) Effect of Temperature (\(\alpha = 0.5\))}
\begin{tabular}{@{}cccc@{}}
\toprule
\textbf{Temperature} & \textbf{Precision} & \textbf{Recall} & \textbf{F1-Score} \\ \midrule
1.0                        & 0.9934              & 0.9933           & 0.9933             \\
2.0                        & 0.9953              & 0.9953           & 0.9953             \\
3.0                        & 0.9942              & 0.9941           & 0.9941             \\ \bottomrule
\end{tabular}

\label{tab:distillation_params}
\vspace{-1em}
\end{table}

To assess the contributions of model distillation and contextual feature embedding, we conduct ablation studies focusing on model complexity reduction and dataset augmentation. Fig. \ref{fig:figure_layer} illustrates the impact of the number of residual layers in the student model. It can be observed that the performance remains largely stable when pruning from $12$ layers (original GPT-2-base) to $3$.   However, a further pruning to a single layer results in a notable accuracy decline, indicating the lower bound of viable model compression. Table \ref{tab:data_augmentation_comparison} summarizes the positive effects of contextual feature embedding, showing consistent improvements across precision, recall, and F1 scores when augmentation is applied.   This enhancement validates that embedding key metadata into input representation bolsters the model’s interpretive capacity, particularly for complex and heterogeneous traffic data.
Table \ref{tab:distillation_params} presents the effect of distillation parameters on classification performance,
% , evaluated using the USTC TFC 2016 dataset with 3-layer model. Subtable (a) examines the impact of varying $\alpha$ while keeping temperature $= 2.0$. Subtable (b) evaluates the effect of temperature with a fixed $\alpha = 0.5$. 
which validates the robustness of MERLOT.

%These ablation results confirm that model distillation significantly reduces computational requirements while preserving accuracy, and that contextual embedding enhances classification robustness, especially in encrypted traffic scenarios.
\begin{remark}
    This consistent performance demonstrates our method’s robustness, particularly in handling encrypted and obfuscated traffic, where specialized models are essential for high classification accuracy. The integration of model distillation and MoE architectures enables our framework to adapt to specific traffic patterns, leading to enhanced precision across diverse classification tasks.
\end{remark}
%\subsection{Discussions}
%To demonstrate the efficiency of our framework, we analyze its superior computational efficiency over TrafficLLM, which is based on the LLaMA2-7B model. Our evaluation focuses on parameter count and computational complexity, two critical factors for deployment in resource-constrained environments.

\section{Conclusion}
\label{sec:conclusion}
In this work, we have introduced MERLOT, a scalable MoE based refinement of distilled LLM for encrypted traffic classification. Through extensive evaluation on diverse datasets, our method has demonstrated competitive or superior performance compared to the state-of-the-art models, particularly in handling encrypted and heterogeneous traffic, but need significantly reduced parameter count and inference time. Ablation studies have further validated the benefits of individual techniques like contextual feature embedding and model pruning. With its significantly reduced parameter count and inference time, MERLOT has proven highly qualified for deployment in resource-constrained, real-time edge environments.
% \bibliographystyle{IEEEtran}
% \bibliography{reference}

\end{document}